\documentclass[runningheads]{llncs}
\usepackage[T1]{fontenc}
\usepackage{cite}
\usepackage{amsmath,amssymb,amsfonts}
\usepackage{comment}
\usepackage{algorithmic}
\usepackage{graphicx}
\usepackage{textcomp}
\usepackage{orcidlink}
\usepackage{marvosym}
\usepackage{pdfpages}
\usepackage{multicol}
\usepackage{booktabs}
\usepackage{multirow}
\usepackage{bigstrut}
\usepackage{xspace}
\usepackage{xcolor}
\usepackage{xspace}

\begin{document}
\title{An Uncertainty-guided Tiered Self-training Framework for Active Source-free Domain Adaptation in Prostate Segmentation
}
\titlerunning{UGTST for Active Source-free Domain Adaptation}

\author{Zihao Luo\inst{1,3}
\and{Xiangde Luo\inst{1,2}}
\and{Zijun Gao\inst{4}}
\and{Guotai Wang\inst{1,2}}}

\authorrunning{Z. Luo et al.}
\institute{
$^1$School of Mechanical and Electrical Engineering, University of Electronic Science and Technology of China, Chengdu, China\\
$^2$Shanghai AI Lab, Shanghai, China\\
$^3$School of Mathematical Sciences, Harbin Engineering University, Harbin, China.\\
$^4$Department of Computer Science and Engineering, The Chinese University of Hong Kong, Sha Tin, Hong Kong.\\
\email{guotai.wang@uestc.edu.cn}\\
}
\maketitle

\footnotetext{Z. Luo and X. Luo contributed equally to this work.}
\begin{abstract}
Deep learning models have exhibited remarkable efficacy in accurately delineating the prostate for diagnosis and treatment of prostate diseases, but challenges persist in achieving robust generalization across different medical centers. Source-free Domain Adaptation (SFDA) is a promising technique to adapt deep segmentation models to address privacy and security concerns while reducing domain shifts between source and target domains. However, recent literature indicates that the performance of SFDA remains far from satisfactory due to unpredictable domain gaps. Annotating a few target domain samples is acceptable, as it can lead to significant performance improvement with a low annotation cost. Nevertheless, due to extremely limited annotation budgets, careful consideration is needed in selecting samples for annotation. Inspired by this, our goal is to develop Active Source-free Domain Adaptation (ASFDA) for medical image segmentation. Specifically, we propose a novel \textbf{U}ncertainty-\textbf{g}uided \textbf{T}iered \textbf{S}elf-\textbf{t}raining (UGTST) framework, consisting of efficient active sample selection via entropy-based primary local peak filtering to aggregate global uncertainty and diversity-aware redundancy filter, coupled with a tiered self-learning strategy, achieves stable domain adaptation. Experimental results on cross-center prostate MRI segmentation datasets revealed that our method yielded marked advancements, with a mere 5\% annotation, exhibiting an average Dice score enhancement of 9.78\% and 7.58\% in two target domains compared with state-of-the-art methods, on par with fully supervised learning. Code is available at: \url{https://github.com/HiLab-git/UGTST}.
\end{abstract}
\section{Introduction}
Automatic and accurate delineation of the prostate plays an important role in assisting the diagnosis and treatment of prostate diseases. Despite that deep learning models have achieved remarkable performance on this task\cite{liu2020ms,jia2019hd}, they often struggle to generalize well when confronted with gaps between training and testing data\cite{luo2023deep}. To tackle this issue, Domain Adaptation (DA) methods emerge as a promising solution\cite{chen2019synergistic}. Unsupervised Domain Adaptation (UDA) has demonstrated considerable efficacy by leveraging knowledge from labeled source domain data to facilitate segmentation on unlabeled target domain\cite{wu2024fpl+,xu2023novel}. Moreover, given the constraints posed by privacy and security concerns, the unavailability of source domain necessitates extensive exploration of Source-Free Domain Adaptation (SFDA) techniques in medical image segmentation\cite{liu2021adapting,chen2021source,wu2023uplaggregate,yang2022source}. Nonetheless, owing to the unforeseeable domain discrepancies, both UDA and SFDA face challenges in achieving satisfactory results.

Recently, a few works\cite{liu2022act,basak2023semi} have confirmed that a small amount of labeled images in the target domain can significantly improve the model's generalizability in the Semi-supervised Domain Adaptation (SSDA) scenario. Despite its performance, SSDA still requires a considerable amount of annotations for DA and still needs to access the source domain. In addition, SSDA overlooks the strategic selection of annotated samples and uses random sample selection with a given annotation budget, which may not select the most valuable images for annotation, leading to sub-optimal performance. In this work, we explore using active learning strategies for effectively selecting valuable samples for annotation\cite{budd2021survey}, which is promising to further reduce the annotation cost, leading to active SFDA (ASFDA). Presently, there is widespread exploration of active sample selection methods grounded in uncertainty-guided approaches\cite{gal2017deep,he2019towards}, feature space diversity\cite{sener2017active}, and their amalgamation\cite{kothandaraman2023salad}. However, due to the complex and dense nature of inherent predictions, along with domain gaps leading to unreliable model features or predictions, conventional active learning methods are unsuitable for ASFDA scenarios. Moreover, as active samples are commonly assumed to harbor the most informative and representative data, they ideally should play a dominant role in the training process. However, this aspect has been neglected by current methods\cite{wu2023upl,yang2022source,wang2023dual, wang2024advancing}.

To mitigate the aforementioned limitations, we propose a practical active learning method Uncertainty-guided Tiered Self-training (UGTST), tailored for ASFDA scenarios in medical image segmentation. In contrast to traditional active learning methods, which often require multiple rounds and utilize only annotated active samples, our approach involves just one round of inference by the source model on the target domain and utilizes unlabeled data in adaptation. We proposed a novel entropy-based slice-level uncertainty estimation method termed global aleatoric uncertainty aggregation and incorporated a diversity-aware redundancy filter for the active sample selection. In response to active samples being undervalued, we developed a Tiered Self-training (TST) DA strategy, by obtaining assumed stable sets to cooperate with active sample dominated DA. 

The contributions of this work can be summarized as follows: (1)We present a novel and efficient ASFDA framework called UGTST for prostate segmentation tasks, aiming to improve target domain generalizability through efficient annotation efforts manageable in clinical practice. (2)A global uncertainty estimation method for active sample selection in medical image segmentation is designed, along with a diversity-aware redundancy filter to achieve stable and efficient active sample selection. (3)We proposed a practical DA strategy TST  for ASFDA, ensuring dominant learning of active samples while progressively utilizing pseudo-labels of unlabeled images. Our method has achieved better performance on the prostate segmentation task than existing ASFDA approaches and was comparable to fully supervised learning with 5\% annotation costs.

\begin{figure}[t]
    \centering
\centerline{\includegraphics[width=12cm]{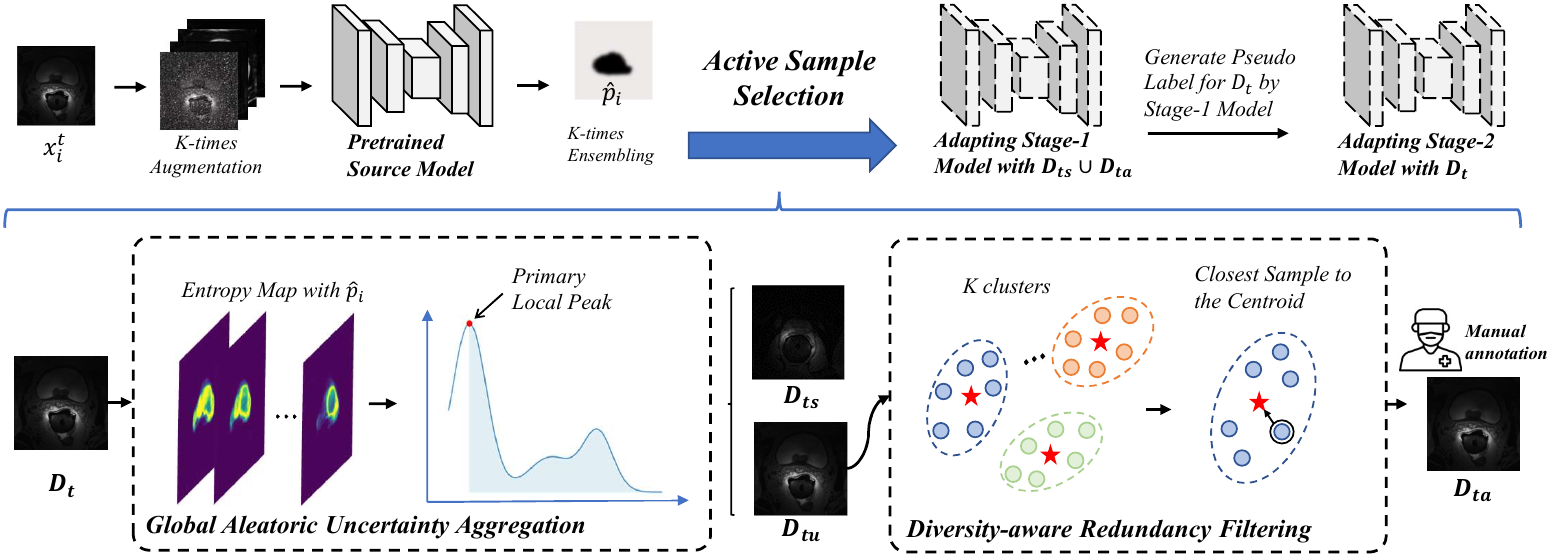}}
    \caption{Overview of our Uncertainty-guided Tiered Self-training Framework, where the $D_t$, $D_{tu}$, $D_{ts}$ and $D_{ta}$ are the target domain set, uncertainty candidate set, assumed stable set, and active sample set, respectively. Our method uses the augmentation-based perturbations output for active sample selection via uncertainty and diversity, then employs a tiered self-training strategy for domain adaptation.}
    \label{fig:method}
\end{figure}
\section{Method}
We consider a scenario where a segmentation model trained on a source domain dataset is deployed to a target domain dataset $D_t = \{(x_i^t)\}_{i=1}^{N_t}$, where $D_t$ is unlabeled at the beginning. The objective of ASFDA is, under a controllable small labeling budget $M$ ($M \ll N_t$), to select a labeled subset of samples $D_{at} = \{(x_i^t, y_i^t)\}_{i=1}^M$ for one round, and utilize $D_t$ to adjust the pre-trained source model $\mathbf{M}_s$ to achieve good dense predictions on the target domain. Our proposed UGTST is depicted in Fig.\ref{fig:method}, for the active sample selection stage, given a labeling budget $M$, we partition the target domain set $D_{t}$ into an uncertainty candidate set $D_{tu}$ and an assumed stable set $D_{ts}$ based on global aleatoric uncertainty aggregation in entropy map. To ensure the diversity of active samples, we further select the active sample set $D_{ta}$ from the uncertainty candidate set $D_{tu}$ through diversity-aware redundancy filtering. Then, a tiered self-training strategy was employed for adaptation.
\subsection{Active Sample Selection via Uncertainty and Diversity}
To highlight the most valuable and informative samples, the entropy-based uncertainty estimation method is a common approach in active learning\cite{budd2021survey}. However, the source model's limited generalizability leads to highly confident yet unstable predictions on the target domain, making direct computation of entropy maps unreliable. To address this, we adopt a test-time augmentation approach, combining predictions with perturbations from diverse augmentation\cite{gaillochet2022taal,wang2019aleatoric}, to diminish confidence in unstable regions and yield more stable predictions\cite{luo2022semi,fu2023cam}. For $x_t^i \in  D_t$, we design intensity augmentation $\mathcal{I}$ and spatial augmentation $\mathcal{T}$, with $K$-times random perturbation, the ensemble segmentation result of $x_t^i$ is:
\begin{equation}
    \hat{p}^i = \frac{1}{K}\sum_{k=1}^K(\mathcal{T}_k^{-1} \circ \mathbf{M}_S\big(\mathcal{I}_k(\mathcal{T}_k \circ x_t^i)))
    \label{con:esem}
\end{equation}
where $\mathcal{I}_k$, $\mathcal{T}_k$ is $k$-th random intensity and spatial transformation and $\mathcal{T}_k^{-1}$ is the corresponding inverse spatial transformation. And for $\hat{p}^i \in \mathbb{R}^{C \times H \times W}$ of $x_t^i$, the entropy map $\boldsymbol{H}(\hat{p}^i) \in \mathbb{R}^{H \times W}$ is calculated as:
\begin{equation}
    \boldsymbol{H}(\hat{p}^i) = -\sum_{c=1}^{C} \hat{p}^i(c) \log(\hat{p}^i(c))
\end{equation}
\\
\textbf{Global Aleatoric Uncertainty Aggregation.}
As mentioned earlier, the entropy map $\boldsymbol{H}(\hat{p}^i)$ cannot be directly used for active sample selection. Due to the imbalance between foreground and background, taking an average of pixel-level uncertainty across the image will be biased to the background. To identify the uncertain region, we design an adaptive threshold to exclude this portion from the output, aiming to aggregate pixels to obtain an unbiased global uncertainty estimation. Hence, we introduce a novel slice-wise uncertainty estimation method called Global Aleatoric Uncertainty Aggregation (GAUA) specifically tailored for medical image segmentation tasks. The discrete density distribution $h^i[n] \in \bar{\boldsymbol{H}}(\hat{p}^i)_{n=1}^{\mathbb{N}}$ is obtained by partitioning the data into bins of size $\mathbb{N}= 100$, arranged from small to large, we can compute the primary local peak value $\boldsymbol{T}_i$ of $x_t^i$ using the discrete difference method, as the self-adaptive threshold to aggregate pixels with relatively high entropy:
\begin{equation}
    \boldsymbol{T}_i = \operatorname*{min}\{h^i[n] | h^i[n]\in\bar{\boldsymbol{H}}(\hat{p}^i)_{n=1}^{\mathbb{N}},|\Delta h^i[n]|<\delta,\Delta^{2}h^i[n]<0\}
\end{equation}
where $\Delta h^i[n]$ is the first-order discrete difference of $h^i[n]$, $\Delta^{2}h^i[n]$ is the second-order one. $\delta$ is a small adaptive bias for approximation. Then, we compute the mean on pixels with relatively high entropy as the GAUA uncertainty $\boldsymbol{U}_{i}$ for $x_t^i$:
\begin{equation}
    \boldsymbol{U}_{i}=\frac{\sum_{n=1}^Nh^i[n] \cdot \mathbb{I}(h^i[n]>\boldsymbol{T}_i)}{\sum_{n=1}^N \mathbb{I} (h^i[n]>\boldsymbol{T}_i)}
\end{equation}
where $\mathbb{I}$ is the indicator function. Then, we divide $D_{t}$ into two parts: 
\begin{equation}
    D_{tu} = \{x_{t}^i | x_{t}^i\in D_{t},\boldsymbol{U}_{i}\geq\boldsymbol{U}^{N_{tu}}_{i}\}; D_{ts} = D_t \setminus D_{tu}
\end{equation}
where $\boldsymbol{U}^{N_{tu}}_i \text{ is the } N_{tu} \text{-th largest value in } \boldsymbol{U}_{i} \text{ corresponding to } D_t$, the capacity $N_{tu}$ of $D_{tu}$ is a hyper-parameter for balancing uncertainty and diversity.\\
\textbf{Diversity-aware Redundancy Filtering.} In the uncertainty candidate set $D_{tu}$, neighboring slices often have similarly high uncertainties. Labeling them would inevitably introduce redundancy, leading to wasted annotation. To deal with this, we take the feature representation $\bar{f}_{x_t^i}$ of slice $x_t^i$ from the encoder of $\mathbf{M}_S$, and we use K-means++\cite{arthur2007k} to cluster $D_{tu}$ into $M$ clusters, which $M$ is the annotation budget, and select the samples closest to the cluster centroids:
\begin{equation}
    D_{ta}=\{\arg\min_{x^{tu}_i\in D_{tu}}||\bar{f}_{x^{tu}_i}-C_k||^2;k=1,2,...,M\}
\end{equation}
where $C_k$ is the centroid of the $k$-th cluster. $||\cdot||^2$ is the Euclidean distance. $\bar{f}_{x^{tu}_i}$ is the feature representation of $x^{tu}_i$. Then, annotators are requested to provide manual annotations for selected samples, leading to an annotated subset $D_{ta}=\{(x^{ta}_i,{y}^{ta}_i)\}_{i=1}^M$.

\subsection{Tiered Self-training for Adaptation}
To mitigate the impact of noisy pseudo-labels on active sample learning and make active samples dominant in training, we propose a Tiered Self-training(TST) strategy. We first train a stage-1 model $\mathbf{M}_{t1}$ initialized with parameters from $\mathbf{M}_S$ on $D_{ta}\cup D_{ts}$, where $D_{ta}$ with labeled samples, $D_{ts}$ with pseudo labels. Then, using the trained $\mathbf{M}_{t1}$, we regenerate pseudo-labels for the unlabeled subset of target domain dataset ${D}_{t} \setminus{D}_{ta} $ with the same strategy in Eq.\ref{con:esem}. Subsequently, we train a stage-2 model $\mathbf{M}_{t2}$ on $D_{t}$, progressively achieving domain adaptation across samples with varying degrees of stability. The average of Dice loss and Cross-Entropy loss is used for self-training.
\section{Experiment and Results}
\subsection{Experimental Details}
\textbf{Dataset.}
To demonstrate the effectiveness of our UGTST method, we employ publicly available prostate T2-weighted MRI images from various clinical centers to evaluate cross-center DA. We select 60 MRI samples comprising a total of 1544 slices from the NCI-ISBI 2013 dataset\cite{bloch2015nci} as the source domain. Additionally, we choose a total of 512 slices from 12 MRI samples acquired from Beth Israel Deaconess Medical Center (BIDMC) and a total of 288 slices from 12 MRI samples obtained from Haokland University Hospital (HK) as two target domains from the PROMISE 12 dataset\cite{litjens2014evaluation}. In the preprocessing stage, we resized all samples to 384×384 in the axial plane and applied min-max normalization to the volume, following previous studies\cite{liu2020ms}. Data from each site were divided into four folds at the case level for cross-validation. We only open the labels of the training set in the target domain during the active sample selection stage, simulating the annotation in clinical practice with a labeling budget of 5\%.\\
\textbf{Implementation Details.}
We tackled the challenge of large inter-slice spacing by employing slice-by-slice segmentation with 2D CNNs, followed by stacking the results into a 3D volume. Our approach utilizes the widely adopted classic 2D U-Net segmentation network\cite{ronneberger2015u}, with its encoder and decoder serving as the feature extractor and prediction head, respectively. Experiments were conducted using PyTorch on an NVIDIA RTX 2080Ti GPU. For the source model, we trained a segmentation network on annotated source data with a batch size of 24 for 400 epochs, using SGD optimization with an initial learning rate of 0.01 and polynomial decay with a power of 0.9. During the adaptation phase, training was conducted for 100 epochs with a batch size of 24, using the same SGD with an initial learning rate of 0.001. We used data augmentation including random spatial transformations (flips and rotations) and intensity transformations (gamma correction, contrast enhancement, Gaussian blur and noise) during training. Dice Similarity Coefficient (DSC) and 95\% Hausdorff Distance ($HD_{95}$) were used as quantitative evaluation metrics in 3D volumes.\\
\begin{table}[t]
  \centering
  \setlength{\tabcolsep}{0.25mm}
  \caption{Quantitative comparison of different domain adaptation methods on prostate segmentation. The best results are in bold, and the second-best are underlined. $^*$ indicates p-value $<$ 0.01, and $^\dag$ (p-value $<$ 0.05) (paired t-test) compared to the second-best.}
  \label{tab:sota}
  \scalebox{1}{
    \begin{tabular}{cc|cc|cc}
    \hline
    \multirow{2}[4]{*}{Task} & \multirow{2}[4]{*}{Method}& \multicolumn{2}{c|}{Target Domain BIDMC} & \multicolumn{2}{c}{Target Domain HK} \\
    \cline{3-6}          & &~~DSC(\%)↑~~& ~$HD_{95}$(mm)↓~  &~~DSC(\%)↑~~& ~$HD_{95}$(mm)↓~  \\
    \hline
     No DA &  Source only  & 45.08±32.63 & 44.20±63.30  & 42.00±30.26 & 21.90±24.51 \\
      No DA & Target-only & 80.59±9.27 & 7.94±8.24 & 81.21±8.10 & 4.14±3.07\\
     Fully DA & Fine-tune & 84.28±4.29 & 5.24±1.58 & 84.83±5.27 & 2.85±0.74 
\\
    \hline
    \multirow{3}[0]{*}{SFDA} & DPL\cite{chen2021source}~ & 67.17±16.03 & 9.26±6.63 & 64.55±18.04 & 6.89±4.94  \bigstrut[t]\\
          & FSM\cite{yang2022source}~ & 72.17±11.21 & 6.05±2.77 & 72.83±12.72 & 4.59±1.32 \\
          & UPL\cite{wu2023upl}~ & 70.21±12.64 & 6.21±3.74 & \underline{73.59±11.51} & \underline{4.36±1.93}  
\\
    \hline
    \multirow{6}[0]{*}{ASFDA} & Random & 65.14±18.23 & 8.17±5.04 & 60.97±26.59 & 11.50±16.42 \\ 
        & CTC\cite{nguyen2004active} & 62.14±21.07 & 10.14±6.87 & 64.60±23.92 & 10.02±13.92 \\ 
        & LC\cite{he2019towards} & \underline{73.68±11.10} & \underline{5.89±1.94} & 69.61±13.93 & 9.11±7.03 \\ 
        & Core-set\cite{sener2017active} & 70.07±15.30 & 6.25±2.93 & 72.12±11.33 & 5.29±1.49 \\ 
        & SALAD\cite{kothandaraman2023salad} & 73.22±11.85 & 6.02±1.63 & 71.48±12.52 & 5.42±2.07 \\
        & \textbf{UGTST(Ours)}  & \textbf{83.46±4.39}$^*$ & \textbf{5.16±1.73} & \textbf{81.17±7.65}$^*$ & \textbf{3.37±1.15}$^\dag$ \\
    \hline
    \end{tabular}%
    }
    \vspace{-0.15cm}
\end{table}%

\begin{figure}[t]
	\centering
	\centerline{\includegraphics[width=12cm]{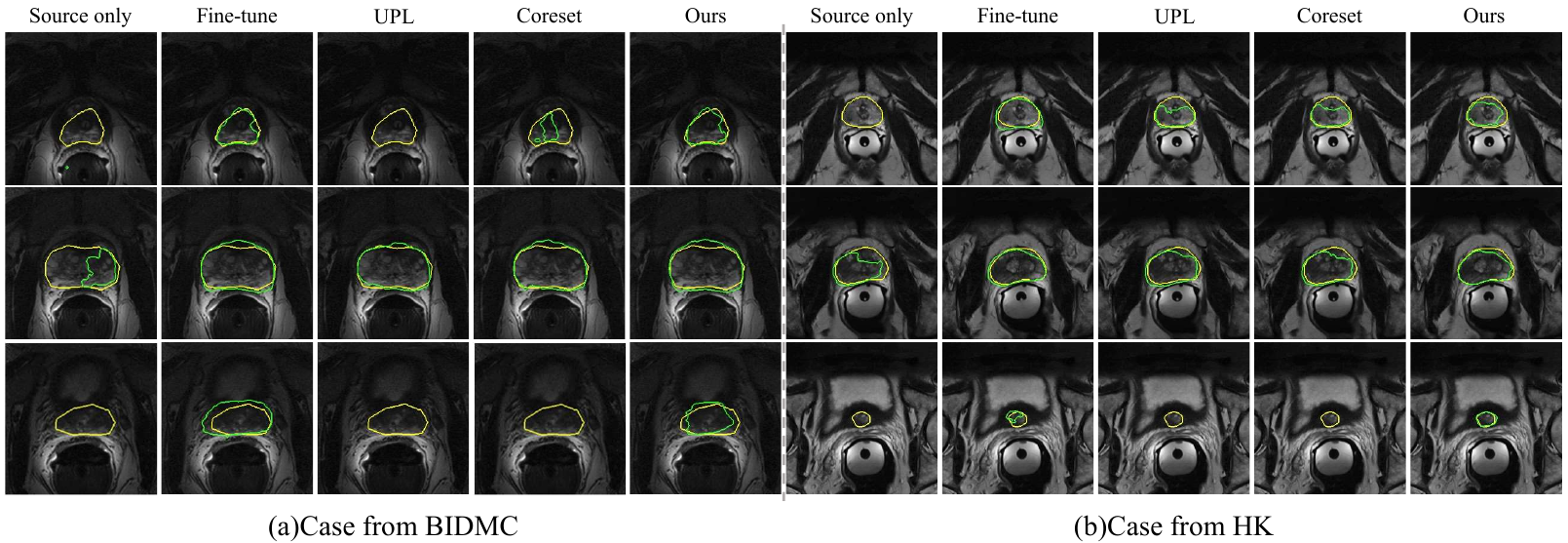}}
   \caption{Qualitative comparison of different Domain Adaptation methods. The ground truth and prediction are displayed in yellow and green contours respectively.
  } \label{fig:segmentation}
\end{figure}

\subsection{Comparison with State-of-the-art Methods.}

Firstly, we investigated the performance of three state-of-the-art SFDA methods: 1)\textbf{DPL}\cite{chen2021source}, 2)\textbf{FSM}\cite{yang2022source}, and 3)\textbf{UPL}\cite{wu2023upl}. Next, our method was compared with five other sample selection methods for annotation with the same budget: 1)\textbf{Random}: randomly select the samples, 2)\textbf{CTC}\cite{nguyen2004active}: select the samples closest to the cluster centers, 3)\textbf{LC}\cite{he2019towards}: samples with smallest probability, 4)\textbf{Coreset}\cite{sener2017active} samples selected by a set-cover problem and 5)\textbf{SALAD}\cite{kothandaraman2023salad}: an ASFDA method employing active learning strategy and guided attention transfer network. These methods were also compared: 1)\textbf{Source only}: The pre-trained source model, serving as the lower bound. 2)\textbf{Target only}: The model was trained solely with annotated images from the target domain. 3)\textbf{Fine-tune}: finetuning the source model with full annotations of the target dataset, serving as the upper bound. For a fair comparison, all methods utilized the same backbone architecture\cite{ronneberger2015u} with post-processing by retaining the largest connected component in a 3D volume.

The quantitative results based on 4-fold cross-validation of adaptation in two target domains are shown in Table \ref{tab:sota}. ``Source only'' and ``Target only'' achieved an average Dice of 45.08\% and 80.59\%, respectively in BIDMC domain and 42.00\% and 81.21\% for HK domain. In observation, SFDA methods demonstrate an enhancement in performance compared to the ``Source only'', FSM\cite{yang2022source} and UPL\cite{wu2023upl} respectively achieved results of 72.17\% and 73.59\% average DSC as the best SFDA method. However, there still exists a considerable gap from the upper bound, underscoring the necessity of ASFDA. In the ASFDA with 5\% labeled data, Random selection achieved an average DSC of 65.14\% and 60.97\%, the corresponding values for the best existing method were 73.68\% and 72.12\%, respectively. Our method achieved DSC of 83.46\% and 81.17\%, significantly improving performance, and achieved comparable results with an upper bound with ``Fine-tune''. Fig.\ref{fig:segmentation} shows qualitative results between different methods in both two target domains. In the central region where the prostate boundary is prominent, most methods show considerable improvement than ``Source only''. However, due to the effective integration of uncertainty and diversity, only our approach achieves high-accuracy segmentation of the prostate region in areas where the boundary is less distinct like the apex and base of the prostate. 
\begin{figure}[t]
	\centering
	\centerline{\includegraphics[width=12cm]{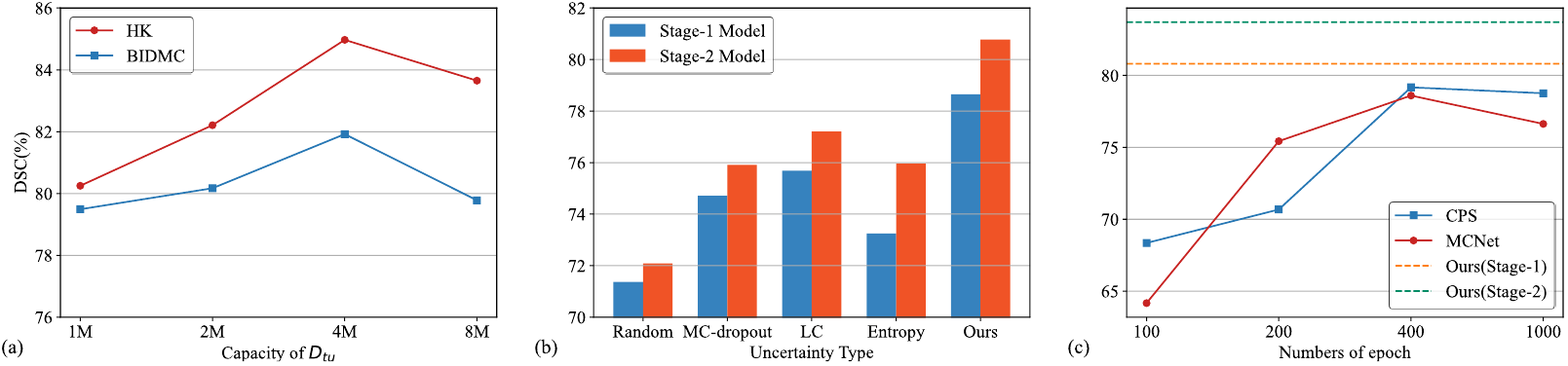}}
	\caption{Ablation study on the validation set. (a). The effect of capacity of $D_{tu}$ with $M=5\%$, (b). Comparison between different uncertainty estimation methods with two-stage results, and (c). Comparison of semi-supervised learning and our method.}
	\label{fig:selection_of_k_t}
\end{figure}
\subsection{Ablation Study.}
To further investigate each component's contribution, we conducted ablation and sensitivity study on the first fold. The capacity $N_{tu}$ of the uncertainty candidate set during active sample selection is a hyper-parameter of our method. We set it to $M$, $2M$, $4M$, and $8M$ to investigate how it affects the performance, where $M = 5\%$, in Fig.\ref{fig:selection_of_k_t}(a). The results from one fold of cross-validation in both domains show that $4M$ is the best hyper-parameter to trade off performance and computational overhead. Further, in Fig.\ref{fig:selection_of_k_t}(b), we also validate the effectiveness of our GAUA compared to other uncertainty estimation methods, including random, MC-dropout\cite{gal2017deep}, Least Confidence(LC)\cite{he2019towards} and highest entropy(Entropy)\cite{wang2014new} followed typical practice of averaging the uncertainty across all the pixels to obtain image-level uncertainty. Our GAUA has achieved the highest performance, and all the methods' performance have been buffed from TST. To demonstrate the necessity of utilizing the source model in the adaptation stage, we employed a few semi-supervised learning methods\cite{chen2021semi,wu2021semi}, using 5\% annotated data selected by our active sample selection technique for Semi-supervised Learning (SSL) in the HK domain under different training epochs. The results are presented in Fig.\ref{fig:selection_of_k_t}(c). The performance of both stages of UGTST surpasses the existing SSL methods, demonstrating the priority and efficiency of DA in ASFDA.

Next, we further validated the contribution of each component of our method in the domain BIDMC. The baseline involved using the source model’s predictions as pseudo-labels for adaptation. ``Augmentation'' means using the ensemble prediction as the pseudo-label to apply self-training process without annotation. When not using TST, we directly merge active samples with labels and unlabeled samples with pseudo labels for self-training. Experimental results shown in Table \ref{tab:ablation} show marked performance improvements by each component of UGTST, further confirming the effectiveness of our approach.\\   

\begin{table}[t]
  \centering
  \setlength{\tabcolsep}{0.25mm}
  \caption{Ablation study of the proposed UGTST method. BIDMC was used as the target domain. The baseline was using the pre-trained model's predictions as pseudo labels for adaptation. GAUA: Global Aleatoric Uncertainty Aggregation, DARF: Diversity-aware Redundancy Filtering, TST: Tiered Self-training.
  }
    \scalebox{1}{\begin{tabular}{cccc|c|c}
    \hline
          & \multicolumn{3}{c|}{Components} & \multirow{2}[4]{*}{DSC(\%)} & \multirow{2}[4]{*}{$HD_{95}$(mm)} \bigstrut\\
\cline{1-4}    Augmentation ~ & GAUA ~ & DARF ~ & TST ~     &       &  \bigstrut\\
    \hline
          ~ & ~ & ~ & ~ & 49.08±16.77 & 8.55±3.67 \bigstrut[t]\\
    \checkmark & ~ & ~ & ~ & 63.21±5.25 & 6.43±1.74 \\
    \checkmark & \checkmark & ~ & ~ & 73.79±2.77 & 5.23±1.15 \\
    \checkmark & \checkmark & \checkmark & ~ & 77.12±5.78 & 4.67±1.25 \\
     \checkmark & \checkmark & \checkmark & \checkmark & \textbf{80.78±4.43} & \textbf{4.37±0.97} \\
    \hline
    \end{tabular}%
    }
  \label{tab:ablation}%
  \vspace{-0.15cm}
\end{table}%

\section{Conclusion}
This work presented an ASFDA framework for accurate prostate segmentation. In the absence of source domain data, active samples are selected by relying on only one round of predictions from a pre-trained source model on the target domain. We present a novel uncertainty-based active sample selection method in medical image segmentation tasks. It utilizes entropy-based primary local peak filtering to aggregate global uncertainty, along with diversity-aware redundancy filters, thus selecting both informative and representative samples for annotation. Then we designed the tiered self-training DA strategy, stabilizing the active learning while progressively leveraging pseudo labels. Our experimental results show that our method achieves comparable performance to fully supervised training with an annotation budget of 5\%, which is manageable in clinical practice. 
\subsubsection{\ackname}
This work was supported by the National Natural Science Foundation of China under grant 62271115.
\subsubsection{\discintname}
The authors have no competing interests to declare that are relevant to the content of this article.
\bibliographystyle{splncs04}
\bibliography{Paper-1781.bib}

\begin{thebibliography}{10}
\providecommand{\url}[1]{\texttt{#1}}
\providecommand{\urlprefix}{URL }
\providecommand{\doi}[1]{https://doi.org/#1}

\bibitem{arthur2007k}
Arthur, D., Vassilvitskii, S., et~al.: k-means++: The advantages of careful seeding. In: Soda. vol.~7, pp. 1027--1035 (2007)

\bibitem{basak2023semi}
Basak, H., Yin, Z.: Semi-supervised domain adaptive medical image segmentation through consistency regularized disentangled contrastive learning. In: MICCAI. pp. 260--270. Springer (2023)

\bibitem{bloch2015nci}
Bloch, N., Madabhushi, A., Huisman, H., Freymann, J., Kirby, J., Grauer, M., Enquobahrie, A., Jaffe, C., Clarke, L., Farahani, K.: Nci-isbi 2013 challenge: Automated segmentation of prostate structures. The Cancer Imaging Archive  (2015), \url{http://doi.org/10.7937/K9/TCIA.2015.zF0vlOPv}

\bibitem{budd2021survey}
Budd, S., Robinson, E.C., Kainz, B.: A survey on active learning and human-in-the-loop deep learning for medical image analysis. MedIA  \textbf{71},  102062 (2021)

\bibitem{chen2019synergistic}
Chen, C., Dou, Q., Chen, H., Qin, J., Heng, P.A.: Synergistic image and feature adaptation: Towards cross-modality domain adaptation for medical image segmentation. In: AAAI. vol.~33, pp. 865--872 (2019)

\bibitem{chen2021source}
Chen, C., Liu, Q., Jin, Y., Dou, Q., Heng, P.A.: Source-free domain adaptive fundus image segmentation with denoised pseudo-labeling. In: MICCAI. pp. 225--235. Springer (2021)

\bibitem{chen2021semi}
Chen, X., Yuan, Y., Zeng, G., Wang, J.: Semi-supervised semantic segmentation with cross pseudo supervision. In: CVPR. pp. 2613--2622 (2021)

\bibitem{fu2023cam}
Fu, J., Lu, T., Zhang, S., Wang, G.: Um-cam: Uncertainty-weighted multi-resolution class activation maps for weakly-supervised fetal brain segmentation. In: MICCAI. pp. 315--324. Springer (2023)

\bibitem{gaillochet2022taal}
Gaillochet, M., Desrosiers, C., Lombaert, H.: Taal: Test-time augmentation for active learning in medical image segmentation. In: MICCAI Workshop on Data Augmentation, Labelling, and Imperfections. pp. 43--53. Springer (2022)

\bibitem{gal2017deep}
Gal, Y., Islam, R., Ghahramani, Z.: Deep bayesian active learning with image data. In: ICML. pp. 1183--1192. PMLR (2017)

\bibitem{he2019towards}
He, T., Jin, X., Ding, G., Yi, L., Yan, C.: Towards better uncertainty sampling: Active learning with multiple views for deep convolutional neural network. In: ICME. pp. 1360--1365. IEEE (2019)

\bibitem{jia2019hd}
Jia, H., Song, Y., Huang, H., Cai, W., Xia, Y.: Hd-net: hybrid discriminative network for prostate segmentation in mr images. In: MICCAI. pp. 110--118. Springer (2019)

\bibitem{kothandaraman2023salad}
Kothandaraman, D., Shekhar, S., Sancheti, A., Ghuhan, M., Shukla, T., Manocha, D.: Salad: Source-free active label-agnostic domain adaptation for classification, segmentation and detection. In: WACV. pp. 382--391 (2023)

\bibitem{litjens2014evaluation}
Litjens, G., Toth, R., Van De~Ven, W., Hoeks, C., Kerkstra, S., Van~Ginneken, B., Vincent, G., Guillard, G., Birbeck, N., Zhang, J., et~al.: Evaluation of prostate segmentation algorithms for mri: the promise12 challenge. MedIA  \textbf{18}(2),  359--373 (2014)

\bibitem{liu2020ms}
Liu, Q., Dou, Q., Yu, L., Heng, P.A.: Ms-net: multi-site network for improving prostate segmentation with heterogeneous mri data. TMI  \textbf{39}(9),  2713--2724 (2020)

\bibitem{liu2022act}
Liu, X., Xing, F., Shusharina, N., Lim, R., Jay~Kuo, C.C., El~Fakhri, G., Woo, J.: Act: Semi-supervised domain-adaptive medical image segmentation with asymmetric co-training. In: MICCAI. pp. 66--76. Springer (2022)

\bibitem{liu2021adapting}
Liu, X., Xing, F., Yang, C., El~Fakhri, G., Woo, J.: Adapting off-the-shelf source segmenter for target medical image segmentation. In: MICCAI. pp. 549--559. Springer (2021)

\bibitem{luo2023deep}
Luo, X., Liao, W., He, Y., Tang, F., Wu, M., Shen, Y., Huang, H., Song, T., Li, K., Zhang, S., et~al.: Deep learning-based accurate delineation of primary gross tumor volume of nasopharyngeal carcinoma on heterogeneous magnetic resonance imaging: A large-scale and multi-center study. Radiotherapy and Oncology  \textbf{180},  109480 (2023)

\bibitem{luo2022semi}
Luo, X., Wang, G., Liao, W., Chen, J., Song, T., Chen, Y., Zhang, S., Metaxas, D.N., Zhang, S.: Semi-supervised medical image segmentation via uncertainty rectified pyramid consistency. MedIA  \textbf{80},  102517 (2022)

\bibitem{nguyen2004active}
Nguyen, H.T., Smeulders, A.: Active learning using pre-clustering. In: ICML. p.~79 (2004)

\bibitem{ronneberger2015u}
Ronneberger, O., Fischer, P., Brox, T.: U-net: Convolutional networks for biomedical image segmentation. In: MICCAI. pp. 234--241. Springer (2015)

\bibitem{sener2017active}
Sener, O., Savarese, S.: Active learning for convolutional neural networks: A core-set approach. arXiv preprint arXiv:1708.00489  (2017)

\bibitem{wang2014new}
Wang, D., Shang, Y.: A new active labeling method for deep learning. In: IJCNN. pp. 112--119. IEEE (2014)

\bibitem{wang2019aleatoric}
Wang, G., Li, W., Aertsen, M., Deprest, J., Ourselin, S., Vercauteren, T.: Aleatoric uncertainty estimation with test-time augmentation for medical image segmentation with convolutional neural networks. Neurocomputing  \textbf{338},  34--45 (2019)

\bibitem{wang2023dual}
Wang, H., Chen, J., Zhang, S., He, Y., Xu, J., Wu, M., He, J., Liao, W., Luo, X.: Dual-reference source-free active domain adaptation for nasopharyngeal carcinoma tumor segmentation across multiple hospitals. TMI  (2024)

\bibitem{wang2024advancing}
Wang, H., Luo, X., Chen, W., Tang, Q., Xin, M., Wang, Q., Zhu, L.: Advancing uwf-slo vessel segmentation with source-free active domain adaptation and a novel multi-center dataset. arXiv preprint arXiv:2406.13645  (2024)

\bibitem{wu2024fpl+}
Wu, J., Guo, D., Wang, G., Yue, Q., Yu, H., Li, K., Zhang, S.: Fpl+: Filtered pseudo label-based unsupervised cross-modality adaptation for 3d medical image segmentation. TMI  (2024)

\bibitem{wu2023upl}
Wu, J., Wang, G., Gu, R., Lu, T., Chen, Y., Zhu, W., Vercauteren, T., Ourselin, S., Zhang, S.: Upl-sfda: Uncertainty-aware pseudo label guided source-free domain adaptation for medical image segmentation. TMI  \textbf{42}(12),  3932--3943 (2023)

\bibitem{wu2021semi}
Wu, Y., Xu, M., Ge, Z., Cai, J., Zhang, L.: Semi-supervised left atrium segmentation with mutual consistency training. In: MICCAI. pp. 297--306. Springer (2021)

\bibitem{xu2023novel}
Xu, X., Chen, Y., Wu, J., Lu, J., Ye, Y., Huang, Y., Dou, X., Li, K., Wang, G., Zhang, S., et~al.: A novel one-to-multiple unsupervised domain adaptation framework for abdominal organ segmentation. MedIA  \textbf{88},  102873 (2023)

\bibitem{yang2022source}
Yang, C., Guo, X., Chen, Z., Yuan, Y.: Source free domain adaptation for medical image segmentation with fourier style mining. MedIA  \textbf{79},  102457 (2022)

\end{thebibliography}
\end{document}